\documentclass[letterpaper]{article}
\usepackage{natbib,alifeconf}
\usepackage{hyperref}
\usepackage{breakurl}

\title{Artificial Life and the Web: WebAL Comes of Age}
\author{Tim Taylor \\
\mbox{}\\
Faculty of Information Technology, Monash University, Clayton,
Victoria 3800, Austalia \\
tim@tim-taylor.com}

\begin{document}
\maketitle

\begin{abstract}
A brief survey is presented of the first 18 years of web-based Artificial Life
(``WebAL'') research and applications, covering the period
1995--2013. The survey is followed by a short discussion of common
methodologies employed and current technologies relevant to WebAL
research. The paper concludes with a quick look at what the future may
hold for work in this exciting area.
\end{abstract}

\section{Introduction}

Four years ago, in 2010, I clicked a link to watch the new video
for the band \emph{Arcade Fire}'s latest release, \emph{We Used to
Wait}. Five minutes later, I was sure that what I had just witnessed
would change the face of Artificial Life research. This was no
ordinary video, but an interactive, localised, personalised
experience, coded using native HTML5 technologies.
Distinct from the song, it goes by its own name of
\emph{The Wilderness
  Downtown}.\footnote{\burl{http://thewildernessdowntown.com/}. For further
  information, see 
\burl{http://b-reel.com/projects/digital/case/57/the-wilderness-downtown/}.} 

On top of the sheer impressiveness of the tightly integrated audio
track and visuals based upon \emph{Google Street View} images of any
address entered by the user at the start of the experience,
real time animation composited directly over the \emph{Street View}
images and guided by the detection of streets within the view,
together with some deft control of action shifting between different browser
windows, \emph{The Wilderness Downtown} features some A-Life related
technologies such as flocking and procedural content generation. 

It graphically illustrates the potential of the Web as a platform for
A-Life applications, and I felt sure when I first watched it that
within the next year or two we would be seeing a great deal of this
kind of work at Artificial Life conferences. However, that hasn't
happened to quite the degree I was expecting, at least not
yet. \emph{The Wilderness Downtown} gained critical acclaim and was a
Grand Prix Winner at the 2011 Cannes Advertising Awards along with a
host of other awards.\footnote{\burl{http://docubase.mit.edu/project/the-wilderness-downtown/}} 
It would appear that, for the time being, commercial development in
this area is somewhat ahead of academic work. 

The rapid development of the Web, and the availability of an ever
increasing number of sophisticated APIs, web-focused languages and
associated technologies, clearly offers rich potential for developing
novel A-Life related research platforms and applications. Despite this
potential, there are still relatively few people working at the
interface of A-Life and the Web (WebAL). However, this is starting to
change. 

In this paper, I highlight some of the historical roots and early work
in this area, some current work, and possible future directions. This
is by no means a comprehensive review, but rather just a taster for
the breadth, depth, and potential of the field. 

Of particular concern in the following are the new methodologies
enabled by web technologies, and the application areas made possible
by those methodologies. I will also highlight some currently relevant
APIs and technologies, although such things are necessarily rather
transient and will doubtless be modified or replaced in the years that
follow. 

\section{Previous Work}
Although the latest HTML5 APIs and related
technologies\footnote{\burl{http://en.wikipedia.org/wiki/HTML5\#New\_APIs}}
offer the possibility of programming sophisticated web applications
natively, without the need of plugins or proprietary extensions, the
idea of using the Web, or, more generally, the Internet, as a platform
for Artificial Life research dates back much earlier.\footnote{The
  historical roots of distributed artificial evolution may be traced
  back to early work on parallel genetic algorithms, with theoretical
  work starting in the 1960s and implementations in the 1980s---see
  \citep{Cantu:Survey} for a good review. One might also consider the
  field of autonomic computing to be relevant, with its focus on large-scale,
  self-managing distributed I.T. systems in the Internet age
  \citep{Kephart:Autonomic}. However, as the focus is on engineering reliable
  I.T. systems rather than A-Life \emph{per se}, I will not discuss this
  further here.}
I divide the following review into what I have called \emph{WebAL 1.0}
and \emph{WebAL 2.0}, in very loose analogy to the popular uptake 
of the term \emph{Web 2.0} around 2004--5 \citep{OReilly:Web2}.\footnote{However, this was
  a gradual transition of ideas and technologies rather than
  a sharp divide, so I do not wish to place too much weight on this distinction.}

\subsection{WebAL 1.0}
In 1995, Tom Ray proposed building a networked version of his well
known A-Life system \emph{Tierra} \citep{Ray:Proposal}. The idea was
to use the Internet to create a large, complex environment in which
digital organisms could roam and freely evolve. Over a period of 5
years or so, Ray and co-workers used \emph{Network Tierra} to
investigate the evolution of complexity in parallel programs (their
analogy to multicellular organisms). Results were mixed: they
succeeded in producing a human-designed multicellular ancestor with
two differentiated cell types (parallel processes) that survived in a
differentiated state under natural selection, but they failed to
achieve an evolutionary increase in the number of cell types
\citep{Ray:SelectingNaturally}. 

The year 1995 also saw the launch of the web-based artificial life virtual
world \emph{TechnoSphere} \citep{Prophet:Technosphere}. The front-end of
the system was a website where users could design their own creatures
by selecting from a limited range of predesigned body parts. Once
created, the user submitted their creature to the web server, and it was
tagged with the user's email address and a unique ID. Submitted
creatures were released into a 3D virtual world (which was not
rendered live on the website), featuring a fractally generated
landscape with trees existing in a certain band of elevations,
other creatures designed by the authors and other online users, and
ecosystem rules governing the interaction between all of these
components. At key moments during a creature's life, and when
interactions occurred with other creatures, the user would receive an
update by email. For interactions with other creatures, the email
addresses of both authors were shared, so that they could discuss the
interaction. Users could also request ``postcards'' of their 
creatures, which were generated by rendering a scene showing the
creature in its current location. In 1996 the \emph{TechnoSphere} world
reached a peak population of 90,000 creatures. In 1998, work started
on a version with real-time 3D rendering \citep{Prophet:Realtime}.
This was exhibited at a number of art galleries and museums over the
period 1999--2001, although this version ran on a local network of PCs
rather than on the Web.

Another early networked A-Life art project was \emph{Life Spacies},
introduced in 1997 and followed by \emph{Life Spacies II} in 1999
\citep{Mignonneau:CreatingALife}. This was an interaction environment
installed in a museum in Tokyo and connected to a website through
which users from all over the world could design virtual creatures
that would then be introduced into the environment displayed at the
museum. The creatures were specified on the website by a novel
text-to-creature encoding system. A related web-based system,
\emph{Verbarium}, was also introduced in 1999, and allowed users to create
shapes and forms in real-time using the same idea of a text-to-form
encoding and an online interactive text editor \citep{Sommerer:Verbarium}.

Moving from art to computer games, the mid-1990s saw the release, in 1996, of
the A-Life focused game
\emph{Creatures}.\footnote{\burl{http://en.wikipedia.org/wiki/Creatures_(artificial_life_series)}}
The main characters in the game were digital life forms, called \emph{Norns},
that were capable of evolution and lifetime learning, and possessed a
physiology, drives, communication abilities, and other life-like skills.
Although the first version of the game ran on standalone PCs, a
growing online community of players soon started exchanging their
\emph{Norns} via enthusiast websites \citep{Jepsen:Creatures}.
In the following years, two further versions of the game were
released, and 2001 saw the release of \emph{Creatures Docking Station}, an
Internet-based add-on to \emph{Creatures 3} that allowed Norns to
travel between different online worlds.\footnote{Source: Wikipedia
  article in previous footnote.} 

A somewhat different kind of A-Life related game was developed by the British
design group \emph{Soda Creative} in 1998. Their system, \emph{Soda
  Constructor}, was written in Java and employed a 2D physics
engine.\footnote{\burl{http://soda.co.uk/work/sodaconstructor}} It
presented users with an online editor with which they could construct
creatures based upon mass-spring systems with oscillating muscles. 
By mid-2000, the popularity of the game had
soared through ``word of email'', and an online forum enabled users to
share their
creations.\footnote{\burl{http://www.acmi.net.au/soda.htm}}
Soda Creative won an Interactive Arts BAFTA Award in 2001 for their
work.\footnote{\burl{http://awards.bafta.org/award/2001/interactive/interactive-arts}}
In 2002, they teamed up with Queen Mary University London
to develop \emph{Sodarace}, a shared online environment where
users from around the world could pit their creations against each
other in
competitions.\footnote{\burl{http://sodarace.net/},
  \burl{http://soda.co.uk/work/sodarace-online-olympics}} The
development of \emph{Sodarace} was supported 
by the UK's Engineering and Physical Sciences Research Council, and
had a strong public outreach and educational flavour.\footnote{In
  2013, Szerlip and Stanley developed an open-source browser-based
  version of Sodarace, called \emph{IESoR} \citep{Szerlip:sodarace}. It
  features a developmental encoding of creatures suitable for
  evolutionary experiments, and is designed to be an accessible
  platform that other researchers can easily use.}

In 2003, Stanley and colleagues initiated development of the computer
game \emph{NERO}, which allowed users to train a team of in-game agents using a
real-time version of the NEAT architecture \citep{Stanley:NERO}. Once
trained, the team could be put to battle against an opposing team
designed by another (possibly remote) user. Battle mode ran on a
server such that both users could watch the battle while running the
program on separate internet-connected machines.\footnote{\emph{NERO} was
  originally distributed as a binary file running on Mac or Windows. In
2009 work commenced an an open-source version called \emph{OpenNERO}
(\burl{http://nn.cs.utexas.edu/?opennero}).}  

To end this \emph{WebAL 1.0} section I take a brief look at some WebAL
systems from the online virtual world \emph{Second Life}, an
environment that itself straddled the transition period from \emph{Web
  1.0} to \emph{Web 2.0}. The two most notable projects are
\emph{Svarga} and \emph{Terminus}, both of which first came to
prominence in 2006. The first of these, \emph{Svarga}, was an island with a
fully functioning ecosystem comprising a weather system and various
types of plants and animals.\footnote{\burl{http://nwn.blogs.com/nwn/2006/05/god_game.html},
\burl{http://nwn.blogs.com/nwn/2010/03/svarga-returns.html}} Shortly
after the release of \emph{Svarga}, a separate effort was launched by the
\emph{Ecosystem Working Group} and associated with the in-game location
\emph{Terminus}.\footnote{\burl{http://news.nationalgeographic.com/news/2007/03/070308-second-life.html}} 
The group's aim was a develop an open source programming language that would
not only allow developers to freely create their own creatures, but
would also allow the creatures in \emph{Terminus} to interact and
evolve using a shared language. Sadly, it seems that the project ran
into funding and resource problems, and is no longer
available.\footnote{\burl{http://forums-archive.secondlife.com/191/83/133314/1.html}}

\subsection{WebAL 2.0}

An interesting early WebAL project that explored the potential of
distributed computation and native client-side storage was \emph{Pfeiffer},
released in late 2001\footnote{I include \emph{Pfeiffer} in the \emph{WebAL 2.0}
  section because of its emphasis on native web technology.} (and still running
today\footnote{\burl{http://www0.cs.ucl.ac.uk/staff/W.Langdon/pfeiffer.html}})
\citep{Langdon:Pfeiffer}. This was a 
browser-based system that allowed users to evolve 2D patterns described
by L-Systems. A user was presented with a variety of patterns on screen,
and could select those they thought were good and bad, which directly
influenced their evolutionary fitness. The user could
also select patterns to be parents for a new offspring. Surviving
patterns were made persistent on the client-side using cookies. Users
could name their favourite patterns, and save them, in which case they
were not only stored locally but also uploaded to the system's global
server where they would become available to be sent to other
users. \emph{Pfeiffer} therefore implemented distributed web-based
evolution with aesthetic selection

One of the first projects to really embrace the potential of
multi-user collaboration provided by \emph{Web 2.0} technologies was the
web-based evolutionary art system \emph{Picbreeder}
\citep{Secretan:Picbreeder}. This is a collaborative interactive
evolution that allows users not only to evolve their own images online
via the project's website,\footnote{\burl{http://picbreeder.org/}} but
also to continue evolving images produced by other
users. \emph{Picbreeder} thereby allows the evolution of very deep
lineages of evolved pictures, and the collective exploration of a vast
search space of images. 

An example of an online game using evolution based upon the behaviour
of multiple distributed users is \emph{Galactic Arms Race (GAR)}
\citep{Hastings:GAR}.\footnote{\burl{http://gar.eecs.ucf.edu}} This
includes an genetic algorithm that evolves new weapons (based upon
particle systems) according to the users' current playing styles. In
single-player mode, the weapons evolve according to the single user,
but in full multiplayer Internet mode the weapons evolve based upon
the aggregate usage of all players. The end result is the continual
introduction of new in-game content based upon the players' tastes.

The \emph{Picbreeder} system, described above, allows the evolution of 2D
images. In 2011, Clune and colleagues introduced the \emph{EndlessForms}
website for the collaborative interactive evolution of 3D
forms.\footnote{\burl{http://endlessforms.com/}} \emph{EndlessForms},
like \emph{Picbreeder}, is based upon an underlying CPNN
representation of form \citep{Clune:EndlessForms}. 

Also in 2011, a project was launched of a rather different nature to
those discussed above. \emph{OpenWorm} is an ``open science'' project to
develop a detailed 3D dynamic simulation of the nematode \emph{C.\
  elegans} \citep{Palyanov:OpenWorm}. Although the simulation itself
is not web-based, the core team are distributed across the world and
have regular team meetings using web-based collaboration tools. The
project website actively seeks to recruit new members to the team,
including scientists, programmers, artists and
writers.\footnote{\burl{http://www.openworm.org/}} All code, data and
models produced by the project are open-source under the MIT
licence. The project also pursues a crowdfunding approach, seeking
donations via the website, and, in 2014, via a successful
\emph{Kickstarter} campaign that raised over
US\$120,000.\footnote{\burl{https://www.kickstarter.com/projects/openworm/openworm-a-digital-organism-in-your-browser}}  

A novel variety of WebAL was reported by
\cite{Auerbach:InterestingImages}. This work evolved 2D images with a
similar representation to that used in \emph{Picbreeder}. However, the
key difference was that the fitness of each image was determined
automatically rather than by user selection, and the fitness function
included a call to \emph{Google Image
  Search}.\footnote{\burl{http://www.google.com/imghp?sbi=1}} The
rationale was that images of interest to humans would return many
similar hits from the Web, hence, number of returned hits was a
component of the fitness function. 

\cite{Hickinbotham:ALifeZoo} describe work using the \emph{YouShare}
software-as-a-service
infrastructure\footnote{\burl{https://portal.youshare.ac.uk/}} to 
create an online ``ALife Zoo''. They demonstrate the potential of the
system by setting up various well known ALife systems as services,
including \emph{Tierra-as-a-service} and
\emph{Avida-as-a-service}. The system allows software written on
diverse architectures to be run in a consistent framework, and for web
visitors to run and interact with the services for research,
education, and archival purposes. 

Finally, another WebAL system with an educational flavour is
\emph{Ludobots},\footnote{\burl{http://www.uvm.edu/~ludobots/}} 
developed by Bongard and colleagues and launched in 2012.
This is an infrastructure for teaching undergraduate-level evolutionary
robotics using 3D simulations and other tools. The simulations are not
web-based, but the website makes available a series of assignments
that anyone can register to complete. Progress involves not just
successfully completing the assignments, but also web-based peer review of other
students' work. Having completed all assignments, a student is
eligible to collaborate on research projects with other graduates of
the system.

\section{Methodologies and Technologies}

The work summarised in the previous section demonstrates a variety of
ways in which the Web can be used for A-Life research and
applications. Some broad categories of methodology are outlined below
(this is by no means an exhaustive list):

\begin{description}

\item[Distributed computation]
It is becoming increasingly possible to use the Web as a distributed
computation platform. Much of the work surveyed above involves some
aspect of distributed computation. The HTML5 and related APIs such as
\emph{Web Socket}, \emph{Web Workers} and \emph{Web Storage} make it
easier to implement these kinds of distributed computation systems
using native technology. Furthermore, a number of technologies are
currently being developed to allow fast client-side processing at
speeds approaching those of local binaries: Mozilla's
\emph{asm.js},\footnote{\burl{http://en.wikipedia.org/wiki/Asm.js}}
and Google's \emph{Native
  Client},\footnote{\burl{http://en.wikipedia.org/wiki/Google_Native_Client}}
are the most prominent of these. 

\item[Human and hybrid computation]
Closely related to the idea of distributed computation on the Web, and
also a feature of much of the work surveyed above, is the idea of
\emph{human} or \emph{hybrid computation}, where some part of the computation is
performed by human users of the system. A \emph{human based genetic
algorithm} was first proposed by \cite{Kosorukoff:HBGA}, and there is a large
literature on the more general areas of \emph{human computation} and \emph{crowd
creativity} (for good reviews, see \citep{Malone:Harnessing},
\citep{Maher:Design}, \citep{Quinn:Human} and
\citep{Yu:Collective}).\footnote{An interesting recent study that
  conceptualises human decision making and creativity as evolutionary
  computation is described by \cite{Sayama:CollectiveHuman}.} 

\item[Cloud APIs]
The work by \cite{Auerbach:InterestingImages}, described above,
illustrates one way in which Cloud interfaces and APIs (in his case,
\emph{Google Image Search}) may be used as components of computational
intelligence systems. It is not hard to think of many other ways in
which Cloud APIs could be employed to provide enhanced capabilities to
WebAL systems.

\item[Persistent systems]
Most A-Life experiments typically run for a few hours, days, or maybe
weeks on a local machine or compute cluster, data is collected,
results are written up, and no further experimentation is done. A
feature of web-based A-Life systems is that they are persistent and
offer the possibility of on-going runs that last for years (or, in
theory, indefinitely). Furthermore, using client-side processing and
data storage APIs (e.g.\ \emph{Web Workers} and \emph{Web Storage}),
these systems can potentially be massively distributed 
and extended across space as well as time. Systems such as \emph{Pfeiffer}
and \emph{Picbreeder}, discussed above, give some indication of the
potential benefits of web-based experiments, and many other types of
long-term experiment can be imagined.

\item[The Web as a Complex Environment] 
Some of the early papers on WebAL, such as \citep{Ray:Proposal} and
\citep{Langdon:Pfeiffer}, discuss the possibility of A-Life agents
roaming the Internet and evolving in the complex environment that it
provides. Some of the experimental work discussed above shows
aspects of this kind of free-roaming agency, but it seems likely that
this kind of ability could be explored and exploited much more
thoroughly. The \emph{Web Socket} API provides a useful
way in which this agent migration can be implemented natively (albeit
always via the server from one client to another).

\item[Crowdfunding]
While not related to WebAL technology as such, another important way
in which the Web can enhance A-Life is through
\emph{crowdfunding} of research and applications. The \emph{OpenWorm}
project, discussed above, is one example of a research effort that has
succeeded in raising significant funds through a \emph{Kickstarter} campaign
and other crowdfunding efforts.

Steve Grand, author of the \emph{Creatures} game discussed above, also
successfully secured \emph{Kickstarter} funding of nearly US\$57,000
in 2011 to develop a new A-Life powered game, currently still under
development.\footnote{\burl{https://www.kickstarter.com/projects/1508284443/grandroids-real-artificial-life-on-your-pc}} 

Another A-Life veteran, Jeffrey Ventrella, has also recently secured
\emph{Kickstarter} funding of over US\$15,000 for his company 
\emph{Wiggle Planet}\footnote{\burl{https://www.wiggleplanet.com/}} to
develop an augmented reality A-Life
game.\footnote{\burl{https://www.kickstarter.com/projects/1582488758/peck-pecks-journey-a-picture-book-that-spawns-virt}}  

Between them, these three projects have raised nearly US\$200,000 of
funding through \emph{Kickstarter}. These examples demonstrate that it
is possible (although still far from easy) to obtain substantial
funding for A-Life projects via crowdfunding.

\end{description}


\section{Looking Forward}

The preceding sections have looked at ways in which web technologies
and A-Life techniques have been combined in domains as diverse as
collaborative design, human computation, education, outreach,
persistent and long-running experiments, the archiving, sharing,
reproduction, and reuse of scientific experiments and platforms, for
collaborative open science, for art, computer games, crowdfunding, and
more. 

As web technology continues to develop, and particularly with the move
towards native APIs in place of proprietary plugins, the potential for
developing complex web-based A-Life research and applications grows
greater each year.

Whether or not a WebAL project is primarily focused on education or
public outreach, the very nature of the Web means that WebAL research
is inherently open and can reach a wide audience (unless steps are
taken to actively prevent this). As funding councils around the world
place increasing emphasis on the public understanding of science,
WebAL is well placed to play a significant role in the communication
of A-Life research to a wide and diverse audience. Furthermore,
WebAL not only enables wide dissemination of results, but it also
promotes public \emph{engagement} and \emph{participation} with A-Life
research.

Looking back over the research reviewed here, it is clear that great
strides have been made over the last 18 years. However, as web
technology and APIs develop, I have the feeling that current work is
only the tip of the iceberg of what could be possible. 
\emph{The Wilderness Downtown}, itself four years old now, still
remains a great showcase of some of the possibilities of the HTML5
era, and yet there are undoubtedly many other possibilities, some as
yet unimagined. Advances will doubtless be made in all of the areas
outlined in the previous section, and likely in completely different
areas as well. 

It is a truly exciting time to be involved in WebAL research. I
cannot predict what advances and achievements will be made over the
next few years, but I look forward to witnessing what emerges, and eagerly
await a WebAL system that gives me a similar sense of awe as when I first
watched \emph{The Wilderness Downtown}.

%


\footnotesize
\bibliographystyle{apalike}
\bibliography{taylor-webal}

\begin{thebibliography}{}

\bibitem[Auerbach, 2012]{Auerbach:InterestingImages}
Auerbach, J.~E. (2012).
\newblock Automated evolution of interesting images.
\newblock In Adami, C., Bryson, D.~M., Ofria, C., and Pennock, R.~T., editors,
  {\em Proceedings of the Thirteenth International Conference on the Simulation
  and Synthesis of Living Systems (ALIFE 13)}, pages 629--630. MIT Press.

\bibitem[Cant{\'u}-Paz, 1998]{Cantu:Survey}
Cant{\'u}-Paz, E. (1998).
\newblock A survey of parallel genetic algorithms.
\newblock {\em Calculateurs parall\`{e}les, r\'{e}seaux et syst\`{e}mes
  r\'{e}partis}, 10(2):141--171.

\bibitem[Clune and Lipson, 2011]{Clune:EndlessForms}
Clune, J. and Lipson, H. (2011).
\newblock Evolving three-dimensional objects with a generative encoding
  inspired by developmental biology.
\newblock In Lenaerts, T., Giacobini, M., Bersini, H., Bourgine, P., Dorigo,
  M., and Doursat, R., editors, {\em Proceedings of the Eleventh European
  Conference on the Synthesis and Simulation of Living Systems (ECAL 2001)},
  pages 141--148. MIT Press.

\bibitem[Hastings et~al., 2009]{Hastings:GAR}
Hastings, E.~J., Guha, R.~K., and Stanley, K.~O. (2009).
\newblock Evolving content in the galactic arms race video game.
\newblock In {\em Proceedings of the IEEE Symposium on Computational
  Intelligence and Games (CIG09)}, pages 241--248, Piscataway, NJ. IEEE.

\bibitem[Hickinbotham et~al., 2013]{Hickinbotham:ALifeZoo}
Hickinbotham, S., Weeks, M., and Austin, J. (2013).
\newblock The {AL}ife {Z}oo: cross-browser, platform-agnostic hosting of
  artificial life simulations.
\newblock In {\em Proceedings of the Twelfth European Conference on the
  Synthesis and Simulation of Living Systems (ECAL 2013)}, pages 71--78. MIT
  Press.

\bibitem[Jepsen, 1999]{Jepsen:Creatures}
Jepsen, D. (1999).
\newblock They're dead, {J}im: Natural selection goes awry in this depressing
  update.
\newblock In {\em Computer Gaming World}, volume 174, pages 364--366.

\bibitem[Kephart and Chess, 2003]{Kephart:Autonomic}
Kephart, J.~O. and Chess, D.~M. (2003).
\newblock The vision of autonomic computing.
\newblock {\em Computer}, 36(1):41--50.

\bibitem[Kosorukoff, 2001]{Kosorukoff:HBGA}
Kosorukoff, A. (2001).
\newblock Human based genetic algorithm.
\newblock In {\em IEEE International Conference on Systems, Man, and
  Cybernetics}, volume~5, pages 3464--3469. IEEE.

\bibitem[Langdon, 2005]{Langdon:Pfeiffer}
Langdon, W.~B. (2005).
\newblock Pfeiffer: A distributed open-ended evolutionary system.
\newblock In Edmonds, B., Gilbert, N., Gustafson, S., Hales, D., and Krasnogor,
  N., editors, {\em AISB'05: Proceedings of the Joint Symposium on Socially
  Inspired Computing (METAS 2005)}, pages 7--13.
\newblock ISBN 1 902956 48 4.

\bibitem[Maher, 2010]{Maher:Design}
Maher, M.~L. (2010).
\newblock Design creativity research: From the individual to the crowd.
\newblock In Taura, T. and Nagai, Y., editors, {\em Design Creativity 2010
  (Proceedings of the International Conference on Design Creativity)}, pages
  41--47. Springer.

\bibitem[Malone et~al., 2009]{Malone:Harnessing}
Malone, T.~W., Laubacher, R., and Dellarocas, C. (2009).
\newblock Harnessing crowds: Mapping the genome of collective intelligence.
\newblock Technical Report 2009-001, MIT Center for Collective Intelligence.

\bibitem[Mignonneau and Sommerer, 2001]{Mignonneau:CreatingALife}
Mignonneau, L. and Sommerer, C. (2001).
\newblock Creating artificial life for interactive art and entertainment.
\newblock {\em Leonardo}, 34(4):303--307.

\bibitem[O'{R}eilly, 2007]{OReilly:Web2}
O'{R}eilly, T. (2007).
\newblock What is {W}eb 2.0: Design patterns and business models for the next
  generation of softwar.
\newblock {\em Communications and Strategies}, 65(1):17--37.

\bibitem[Palyanov et~al., 2012]{Palyanov:OpenWorm}
Palyanov, A., Khayrulin, S., Larson, S.~D., and Dibert, A. (2012).
\newblock Towards a virtual {C}.\ elegans: A framework for simulation and
  visualization of the neuromuscular system in a {3D} physical environment.
\newblock {\em In {S}ilico Biology}, 11(3):137--147.

\bibitem[Prophet, 1996]{Prophet:Technosphere}
Prophet, J. (1996).
\newblock Sublime ecologies and artistic endeavors: Artificial life and
  interactivity in the online project {T}echno{S}phere.
\newblock {\em Leonardo}, 29(5):339--344.

\bibitem[Prophet, 2001]{Prophet:Realtime}
Prophet, J. (2001).
\newblock Techno{S}phere: ``real'' time, ``artificial'' life.
\newblock {\em Leonardo}, 34(4):309--312.

\bibitem[Quinn and Bederson, 2011]{Quinn:Human}
Quinn, A.~J. and Bederson, B.~B. (2011).
\newblock Human computation: A survey and taxonomy of a growing field.
\newblock In {\em Proceedings of the SIGCHI Conference on Human Factors in
  Computing Systems}, pages 1403--1412. ACM.

\bibitem[Ray, 1995]{Ray:Proposal}
Ray, T.~S. (1995).
\newblock A proposal to create a network-wide biodiversity reserve for digital
  organisms.
\newblock Technical report, ATR, Japan.
\newblock TR-H-133.

\bibitem[Ray, 1998]{Ray:SelectingNaturally}
Ray, T.~S. (1998).
\newblock Selecting naturally for differentiation: Preliminary evolutionary
  results.
\newblock {\em Complexity}, 3(5):25--33.

\bibitem[Sayama and Dionne, 2014]{Sayama:CollectiveHuman}
Sayama, H. and Dionne, S.~D. (2014).
\newblock Studying collective human decision making and creativity with
  evolutionary computation.
\newblock http://arxiv.org/abs/1406.6291.

\bibitem[Secretan et~al., 2008]{Secretan:Picbreeder}
Secretan, J., Beato, N., {D'Ambrosio}, D.~B., Rodriguez, A., Campbell, A., and
  Stanley, K.~O. (2008).
\newblock Picbreeder: evolving pictures collaboratively online.
\newblock In {\em Proceedings of the SIGCHI Conference on Human Factors in
  Computing Systems}, pages 1759--1768. ACM.

\bibitem[Sommerer and Mignonneau, 1999]{Sommerer:Verbarium}
Sommerer, C. and Mignonneau, L. (1999).
\newblock {VERBARIUM} and {LIFE SPACIES}: Creating a visual language by
  transcoding text into form on the internet.
\newblock In {\em IEEE Symposium on Visual Languages (VL'99) Conference
  Proceedings}, pages 90--95, Tokyo.

\bibitem[Stanley et~al., 2005]{Stanley:NERO}
Stanley, K.~O., Bryant, B.~D., and Miikkulainen, R. (2005).
\newblock Evolving neural network agents in the nero video game.
\newblock In {\em Proceedings of the IEEE 2005 Symposium on Computational
  Intelligence and Games (CIG’05)}, pages 182--189, Piscataway, NJ. IEEE.

\bibitem[Szerlip and Stanley, 2013]{Szerlip:sodarace}
Szerlip, P. and Stanley, K. (2013).
\newblock Indirectly encoded {S}odarace for artificial life.
\newblock In Li\`{o}, P., Miglino, O., Nicosia, G., Nolfi, S., and Pavone, M.,
  editors, {\em Proceedings of the Twelfth European Conference on the Synthesis
  and Simulation of Living Systems (ECAL 2013)}, pages 218--225. MIT Press.

\bibitem[Yu et~al., 2012]{Yu:Collective}
Yu, L.~L., Nickerson, J.~V., and Sakamoto, Y. (2012).
\newblock Collective creativity: Where we are and where we might go.
\newblock In Malone, T.~W. and von Ahn, L., editors, {\em Proceedings of
  Collective Intelligence}.
\newblock http://arxiv.org/abs/1204.2991.

\end{thebibliography}

\end{document}